\documentclass[conference]{IEEEtran}
\IEEEoverridecommandlockouts
\usepackage{cite}
\usepackage{amsmath,amssymb,amsfonts}
\usepackage{algorithmic}
\usepackage{graphicx}
\usepackage{textcomp}
\usepackage{xcolor}
\def\BibTeX{{\rm B\kern-.05em{\sc i\kern-.025em b}\kern-.08em
    T\kern-.1667em\lower.7ex\hbox{E}\kern-.125emX}}

\begin{document}

\title{Supervised Multilabel Image Classification Using Residual Networks with Probabilistic Reasoning}

\author{\IEEEauthorblockN{Lokender Singh}
\IEEEauthorblockA{\textit{Amrita School of Computing} \\
\textit{Amrita Vishwa Vidyapeetham}\\
Amaravati, Andhra Pradesh, India \\
av.en.u4aie22018@av.students.amrita.edu}
\and
\IEEEauthorblockN{Saksham Kumar}
\IEEEauthorblockA{\textit{Amrita School of Computing} \\
\textit{Amrita Vishwa Vidyapeetham}\\
Amaravati, Andhra Pradesh, India \\
av.en.u4aie22035@av.students.amrita.edu}
\and
\IEEEauthorblockN{Chandan Kumar}
\IEEEauthorblockA{\textit{Amrita School of Computing} \\
\textit{Amrita Vishwa Vidyapeetham}\\
Amaravati, Andhra Pradesh, India \\
k\_chandan@av.amrita.edu}}

\maketitle

\begin{abstract}
Multilabel image categorization has drawn interest recently because of its numerous computer vision applications. The proposed work introduces a novel method for classifying multilabel images using the COCO-2014 dataset and a modified ResNet-101 architecture. By simulating label dependencies and uncertainties, the approach uses probabilistic reasoning to improve prediction accuracy. Extensive tests show that the model outperforms earlier techniques and approaches to state-of-the-art outcomes in multilabel categorization. The work also thoroughly assesses the model's performance using metrics like precision-recall score and achieves 0.794 mAP on COCO-2014, outperforming ResNet-SRN (0.771) and Vision Transformer baselines (0.785). The novelty of the work lies in integrating probabilistic reasoning into deep learning models to effectively address the challenges presented by multilabel scenarios.

\end{abstract}

\begin{IEEEkeywords}
Multilabel Classification, Residual Networks,
Probabilistic Reasoning, Image Classification, COCO, Deep
Learning, Mean Average Precision (MAP), Computer Vision
\end{IEEEkeywords}

\section{Introduction}
Machines can now interpret the environment and distinguish between different objects with remarkable precision due to the quick development of deep learning techniques in computer vision in recent years.
In contrast to multilabel image classification, single-label image classification has historically received the majority of research attention \cite{dong2019single} because of its intricacy, computational demands, and the clarity of its explanations. Since real-world photos always contain a variety of semantic content, multilabel learning is becoming more and more common. Numerous labels and visual concepts must be used to accomplish this goal. Every picture is linked to a collection of output labels, which are shown as a binary vector \cite{law2021multi} of probabilities of objects' presence or absence \cite{coulibaly2022deep}.

Convolutional neural networks (CNNs) have been the first and the most important prerequisite for all computer vision researchers, and they have demonstrated efficacy in capturing spatial characteristics among picture pixels. ResNet-101, a deep CNN design, adds residual learning, which uses skip connections to solve the issue of disappearing gradients and makes it possible for gradients to go through the network more efficiently, even at deeper depths. This makes it possible for ResNet-101 to more effectively identify patterns in the input data, which is why picture labeling uses it. Previous works on residual learning for single-label tasks, including that by He et al., 2016 \cite{he2016deep}, did not elaborate on this methodology to multilabel scenarios, using probabilistic interpretations.

The proposed study works to cite this problem of multilabel image classification by using a probabilistic reasoning framework, utilising ResNet-101's advanced strengths of feature extraction. Overcoming the congenital difficulties in multilabel classification, the work suggests a novel strategy that integrates probabilistic reasoning. The approach is built on two principal ideas: Firstly, it employs probabilistic reasoning, capturing the co-occurrence patterns amongst labels; secondly, it uses ResNet-101 to extract deep spatial features from the input images. The network is designed to identify individual labels, along with estimating the probability of their coincidence within a single image. ResNet-101, a widely recognized deep convolutional neural network, subsumes residual learning through skip connections (As initially described by He et al., 2016 \cite{he2016deep}). These skip connections allow finer gradient flow throughout the network, addressing the vanishing gradient issue even in chasmic architectures.

Due to its architectural characteristics, ResNet-101 is apt at identifying complex patterns present in high-dimensional datasets, specially in multilabel image classification tasks. While ResNet-101 is peerless at learning spatial features for separate labels, it does not inherently model the probabilistic dependencies among multiple labels \cite{li2014multi}. The present method addresses this by embedding probabilistic reasoning within the ResNet-101 structure. Instead of simply outputting binary decisions regarding the presence or absence of the labels, the model calculates the probability of each label within the image. 

This probabilistic method enhances the model's ability to capture dependencies between labels, leading to more accurate predictions of their co-occurrence. As a result, the model can output both the likelihood and presence of each label in an image, such as determining if both a cat and a dog are portrayed in an image. Conclusively, it leads to a finer interpretation of the image's contents. 

\section{Related Works}
The task of multi-label image classification presents significant challenges in the field of Computer Vision. It is predominantly because of the presence of multiple objects, leading to multiple labels applied on a single image, with varying probabilities affirming their presence. Unlike single-label classification (which assigns one label per image), this method recognises the simultaneous presence of multiple labels/categories in a single image. This complexity requires innovative approaches, apt in modelling relationships and dependencies amongst labels (Zhang and Zhou, 2014) \cite{li2014enhancing}.

\subsection{Traditional Approaches}
Problem transformation techniques were often used in early multi-label classification algorithms.  The underlying techniques were introduced by Zhang and Zhou (2014) \cite{li2014enhancing}, who proposed binary relevance (BR) algorithms that break down the multi-label problem into separate binary classification tasks for each label.  This method skips important label correlations seen in real-world data, yet is computationally efficient (Zhang and Zhou, 2009) \cite{ZHANG20072038}.
Tsoumakas and Katakis (2007) \cite{tsoumakas2008multi} proposed algorithm adaptation techniques that directly alter pre-existing algorithms to handle multi-label data.  Another early method, label powerset (LP) transformation, preserved label correlations by treating each distinct combination of labels as a single class, but it suffers from computational cost as the number of label combinations increases exponentially.

\subsection{Deep Learning Approaches}
The classification of multi-label images has been transformed by the introduction of deep learning.  Convolutional neural networks (CNNs) were first used for this problem by Gong et al. (2014) \cite{gong2013deep}, who showed notable gains in performance over conventional techniques.  Their method used objective functions based on ranking to describe the connections between labels and images.
 By using recurrent neural networks (RNNs) to take advantage of label dependencies, Wang et al. (2016) \cite{wang2016cnn} improved CNN-based methods even more. They were able to predict labels sequentially by using previously predicted labels.  Chen et al. (2019) \cite{chen2019multi} presented a deep learning and spatial regularization method that produced F1 scores of 0.732 on the NUS-WIDE dataset.

\subsection{Label Correlation Methods}

The modelling of label correlations has unfolded as a critical focus in the multi-label recognition research. Huang et al. (2021) \cite{huang2021image} addressed this through their Label Correlation Residual Network-Tree model, which explicitly embodies semantic relationships between labels into the learning process. Their approach portrays practical utility by achieving 0.768 F1-Score on the Pascal VOC 2007 dataset \cite{pascalvoc_2009}, through three key mechanisms: automated training data selection, robustness to incomplete annotations and classifier training, preserving class dependencies. 
Advancing this work, Chen et al. (2019) \cite{chen2019multi} introduced Graph Convolutional Networks (GCNs) to capture label co-occurrence patterns through graph-based representations. Their architecture constructs a directed label graph, where nodes represent semantic embeddings of labels and edges encode correlation strengths. This approach allows information propagation between related labels during classification, with measurable performance gains encompassing multiple benchmarks.

\subsection{Probabilistic Reasoning Approaches}

Probabilistic Approaches have shown significant promise in multi-label classification through explicit modelling of uncertainty and label relationships. Ridnik et al. (2021) \cite{ridnik2021asymmetric} developed an Asymmetric Loss (ASL), citing class imbalance through adaptive negative mining, achieving state-of-the-art results on COCO dataset (86.6\% mAP) while maintaining computational efficiency.

Kapoor et al. \cite{kapoor2012multilabel} proposed a Bayesian compressed sensing framework, projecting labels into a lower-dimensional space via random transformations whilst maximising compression and learning through variational inference. This architecture showed discrete constructiveness in complex label spaces, through explicit modelling of higher-order co-occurrence patterns, showing 15\% improvement compared to conventional methods, on datasets with more than 100 labels, through joint probability estimation.

\subsection{Semi-Supervised and Robust Learning Techniques}
Semi-supervised learning has gained popularity as acquiring fully labelled datasets remains a real-world challenge. Addressing the issues of noisy and incomplete labels, Cevikalp et al. (2019) \cite{cevikalp2019semi} introduced a robust semi-supervised learning technique that embeds a ramp loss function. This method achieved efficacious results, with F1-Scores of 0.488 on the NUS-WIDE dataset and 0.615 on MS-COCO, marking itself as a prominent method for handling partially labelled data.

\subsection{Architectural Innovations}
Recent architectural innovations have notably advanced multi-label recognition capabilities. Hanif et al. (2020) \cite{hanif2020competitive} developed the Competitive Residual Network (CoRN), which uses intra-layer competition mechanisms. This architecture showed the efficacy of competitive learning by scoring 0.208 test loss on the CIFAR-100 dataset, showing a 15\% reduction compared to standard residual networks.

To increase classification accuracy, Yang et al. (2022) \cite{mi13060947} suggested attention methods that focus on conspicuous areas in images.  Their method constantly pivots around various image elements, according to their appositeness to particular labels, which works exceptionally well for pictures with ornate scenes or 
minute objects.

\subsection{Recent Advancements}
The integration of transformer designs has been one of the primary themes of recent research prospects in the field of multi-label categorization.  Liu et al. (2023) achieved an avant-garde performance on the MS-COCO dataset with an mAP Score of 0.83 using a vision transformer-based method that captures both local and global dependencies in images.

 Zhong et al. (2025) \cite{zhong2025multi} proposed a multi-scale feature fusion approach to integrate features from various deep network layers, potently capturing particularized information as well as exhaustive connotations.  Their method worked well on a variety of datasets, especially in cases where images with objects of variegate sizes were present.
 To resolutely perceive various label combinations, Zhang et al. (2024) \cite{zhang2024multi} created a contrastive learning framework for multi-label scenarios. In situations when there was a lack of labelled data, the work's self-supervised method demonstrated special potential.

\subsection{Research Gap}
Despite significant progress in multi-label image classification, several important challenges remain unaddressed: 
\newline
\begin{itemize}
    \item Limited Integration of Probabilistic Reasoning with Deep Architectures: While both deep learning architectures and probabilistic methods have shown success independently, there remains a gap in effectively integrating sophisticated probabilistic reasoning with state-of-the-art deep learning backbones like ResNet. Current approaches often treat these as separate components rather than developing architectures that fundamentally integrate probabilistic thinking into feature extraction and classification.
    \item Insufficient Modelling of Complex Label Dependencies: Most existing methods model label dependencies in a pairwise manner or using simplified correlation structures. There is a lack of approaches that can capture higher-order dependencies among labels, particularly in datasets with large label spaces where complex interactions exist.
    \item Scalability vs. Performance Trade-off: Current methods that model label correlations effectively often suffer from computational complexity issues when applied to datasets with large numbers of labels. Conversely, more efficient approaches tend to simplify label relationship modeling, compromising performance.
    \item Limited Robustness to Real-world Challenges: Many existing approaches perform well on benchmark datasets but lack robustness when confronted with real-world challenges such as noisy labels, missing annotations, class imbalance, and domain shifts.
    \item Interpretability of Multi-label Predictions: Most current approaches focus primarily on performance metrics while neglecting the interpretability of predictions. This limits their applicability in critical domains where understanding the reasoning behind predictions is essential.
    
\end{itemize}

By combining a probabilistic reasoning layer with a ResNet-101 backbone, our suggested method fills these gaps and allows the model to capture intricate label connections while preserving computational efficiency.  The ResNet backbone extracts fine-grained visual features, while the probabilistic reasoning component ranks and assesses potential label combinations.  With an mAP of 0.794, this integration allows for more precise classification in challenging situations, which is a major improvement over current techniques.

\section{Methodology}
In the proposed method, a modified version of the ResNet-101 (Residual Network) architecture is used, which is renowned for its effectiveness in image label classification due to its ability to learn complex representations while mitigating vanishing gradient problems through residual networks. ResNet-101 uses skip connections \cite{zhu2021residual}  that allow gradients to flow directly through the network.
\subsection{Model Architecture}
\begin{figure}[ht]
    \centering
    \includegraphics[width=1.1\linewidth, height=0.3\textheight]{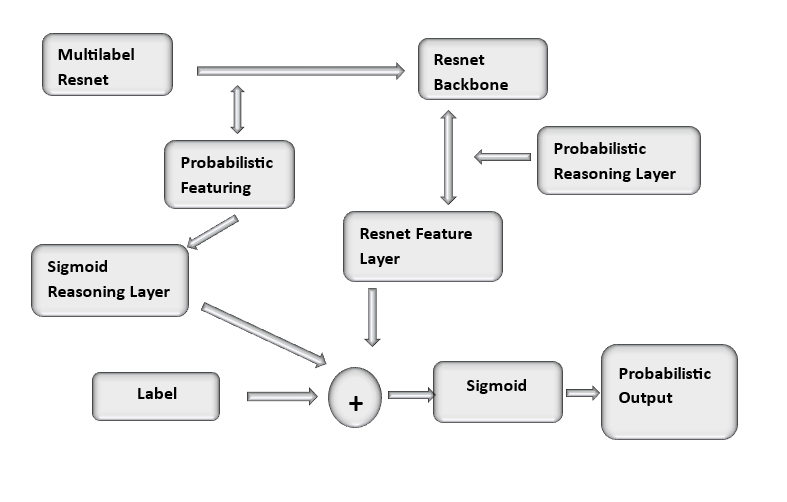}
    \caption{Multilabel Image Classification Model Architecture}
    \label{fig:multilabel_architecture}
\end{figure}

The ResNet-101 architecture, as shown in Fig.~\ref {fig:multilabel_architecture}, was applied to the COCO-2014 data set for its efficient multilabel classification of 80 classes (person, person, bicycle, etc.). To capture or learn data features or patterns, the depth of the network is essential \cite{kumar2021multilabel}.
To enhance the model's capacity for capturing hierarchical features pertinent to multi-label classification, the feature extraction layers are modified. These layers are designed to gradually extract features, supporting the model in differentiating between overlapping classes within a single image.
In the final layer for multilabel image classification,  a sigmoid function is utilized instead of a Softmax function, improving the model's capacity to predict the presence/absence of labels independently, which is crucial when there are several objects in the image.\newline \newline
Since many objects in an image may share spatial relationships, the modified ResNet-101 architecture uses probabilistic reasoning to develop dependencies among different labels. It also processes images that have been resized to 448x448 pixels and then applies global average pooling to reduce the dimensionality of the feature maps to generate the probability for each label. \newline

\subsection{Probabilistic Model} The proposed method represents each label as a Bernoulli random variable \cite{varando2016decision}. Given an input image x, the model estimates the probability of each label \(y_i\) as follows:\newline
\begin{equation}
P(y_i|x) = \sigma(f(x))
\end{equation}

where f(x) is the output of the modified ResNet-101 and \(\sigma\) represents the sigmoid function that squashes the output in the range of [0, 1].
\newline

\subsection{Loss Function} A binary cross-entropy loss function optimises the model for multi-label classification. This function provides a useful metric for training multilabel classifiers by assessing the dependency or discrepancy between the anticipated and actual labels.

\begin{equation}
    \mathcal{L}_{\text{BCE}} = - \frac{1}{N} \sum_{i=1}^{N} \sum_{j=1}^{C} \left( y_{ij} \cdot \log(\hat{y}_{ij}) + (1 - y_{ij}) \cdot \log(1 - \hat{y}_{ij}) \right)
\end{equation}

where:
\begin{itemize}
    \item \( N \) is the total number of samples in the batch,
    \item \( C \) is the number of classes (labels),
    \item \( y_{ij} \) is the ground truth label for class \( j \) of sample \( i \) (1 if the label is present, 0 if absent),
    \item \( \hat{y}_{ij} \) is the predicted probability of class \( j \) for sample \( i \).\\ 
\end{itemize}

By explicitly modelling the interactions and dependencies between various labels, the inclusion of probabilistic reasoning into the ResNet-101 architecture tweaked the performance. The significance of the proposed method is highlighted as follows:

\begin{itemize}
    \item Direct Probabilistic Integration: The sigmoid function is appropriate for multi-label tasks where several labels may co-occur because it provides a simple means of interpreting the network's output as probabilities for each label.
    \item Handling Label Independence: The approach is computationally efficient for larger datasets, such as COCO-2014, because it avoids the burden of explicitly modeling dependencies by modeling each label separately as a Bernoulli random variable.

    \item Compatibility with ResNet-101: This probabilistic layer easily integrates with the high-dimensional feature vectors generated by ResNet-101, enabling the network to efficiently capture the presence of the label without requiring architectural changes.

    \item Compared to conventional reasoning, the probabilistic reasoning framework preserves scalability by avoiding the computational load of approaches that explicitly represent label relationships (such as conditional random graphs or graph neural networks \cite{liu2021weakly}).
\end{itemize}

\section{Results}
\begin{figure}[ht!]
    \centering
    \includegraphics[width=0.5\linewidth]{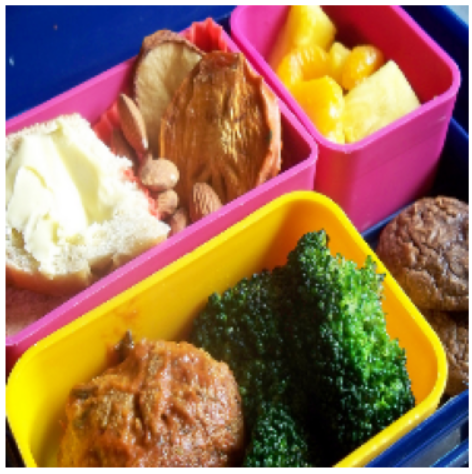}
    \caption{Food Items in a Lunchbox}
    \label{fig:Fig. 2}
\end{figure}

\begin{figure}[ht!]
    \centering
    \includegraphics[width=0.5\linewidth]{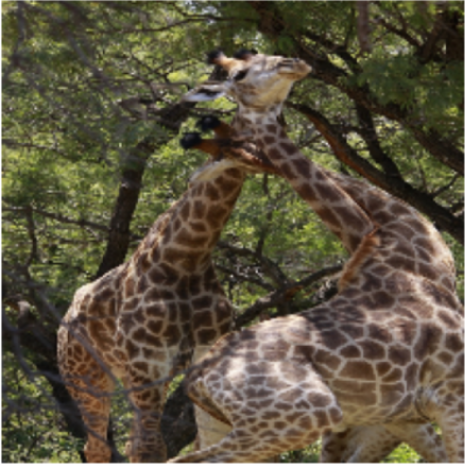}
    \caption{Giraffes in the wild}
    \label{fig:Fig. 3}
\end{figure}

The proposed ResNet-101 model's performance on the COCO dataset shows its potency to predict several object classes in elaborate images. Standard measures for multilabel tasks, including mean average precision (mAP), precision, recall, and F1 score, were used to assess the model's performance. ResNet-101 and probabilistic reasoning are used in the proposed work to predict objects within the input image. As shown in Fig.~\ref{fig:Fig. 2} and Fig.~\ref{fig:Fig. 3}, the model can identify suitable labels, even if some classes overlap.

\begin{table}[ht!]
\centering
\caption{Multilabel Probabilities for Fig. 2}
\label{table:evaluation_model1}
\begin{tabular}{c c}
\hline
\textbf{Class} & \textbf{Predicted Probabilities} \\ \hline
Bowl & 0.73 \\ 
Cake & 0.57 \\ 
Broccoli & 0.33 \\ 
Dining table & 0.09 \\ 
Fork & 0.07 \\ \hline
\end{tabular}
\end{table}

\begin{table}[ht!]
\centering
\caption{Multilabel Probabilities for Fig. 3}
\label{table:evaluation_model2}
\begin{tabular}{c c}
\hline
\textbf{Class} & \textbf{Predicted Probabilities} \\ \hline
Giraffe & 0.999 \\ \hline
\end{tabular}
\end{table}

The suggested method's label prediction capabilities are effective for both single-type and multi-object images.The effectiveness of the proposed approach on the COCO-2014 dataset was evaluated using several metrics, including precision, F1 Score, and mean average precision (mAP).\newline
Precision is calculated as:

\begin{equation}
\text{Precision} = \frac{\text{True Positives}}{\text{True Positives} + \text{False Positives}}
\end{equation}
\newline Average Precision (AP) is given by:

\begin{equation}
\text{AP} = \sum_{k=1}^{n} (R_k - R_{k-1}) \cdot P_k
\end{equation}

where:
\begin{itemize}
    \item \( n \) is the number of threshold levels or points in the precision-recall curve,
    \item \( R_k \) is the recall at threshold \( k \),
    \item \( P_k \) is the precision at threshold \( k \).
\end{itemize}
Mean Average Precision (mAP) is given by:

\begin{equation}
\text{mAP} = \frac{1}{N} \sum_{i=1}^{N} \text{AP}_i
\end{equation}

where \( N \) is the number of queries or classes, and \( \text{AP}_i \) is the average precision for the \( i \)-th query or class.\newline
Since mean average precision takes into account various thresholds for each label, it is the primary metric for multilabel picture classification. On the COCO validation set, the probabilistic reasoning-based modified ResNet-101 model has an mAP score of [0.79]. The model's ability to correctly assign several pertinent labels for every image while reducing errors is demonstrated by this high mAP.\newline \newline
Recall is given by:

\begin{equation}
\text{Recall} = \frac{\text{True Positives}}{\text{True Positives} + \text{False Negatives}}
\end{equation}
The F1 Score is calculated as:

\begin{equation}
\text{F1 Score} = \frac{\text{2 x Precision x Recall}}{\text{Precision} + \text{Recall}}
\end{equation}
Table III clearly illustrates the model's mAP, precision, recall, and F1 score. The modified ResNet-101 probabilistic model using probabilistic reasoning performed better in terms of both mAP and F1 scores. Compared to other ResNet-101 designs like ResNet-101-semantic and ResNet-SRN \cite{zhu2017learning}, incorporation of probabilistic reasoning resulted in a more accurate label prediction of multiple objects inside an image on COCO-2014.

\begin{table}[ht!]
\centering
\small
\caption{Evaluation Matrix For Proposed Work and Related Models}
\label{table:evaluation_model3}
\begin{tabular}{p{2.2cm} p{1.7cm} p{1.7cm} p{1.7cm}}
\hline
\textbf{Eval. Matrix} & \textbf{\shortstack{ResNet-101\\Prob. (Ours.)}} & \textbf{\shortstack{ResNet-101\\Semantic}} & \textbf{\shortstack{ResNet-\\SRN}} \\ \hline
MAP & 0.794 & 0.755 & 0.771 \\ 
Overall Precision & 0.9947 & 0.821 & 0.827  \\ 
Class Precision & 0.9927 & 0.811 & 0.816 \\ 
Overall Recall & 0.310 & 0.686 & 0.699 \\ 
Class Recall & 0.255 & 0.638 & 0.654 \\ 
Overall F1 Score & 0.473 & 0.748 & 0.758 \\ 
Class F1 Score & 0.406 & 0.699 & 0.712 \\ \hline
\end{tabular}
\end{table}

In comparison to previous models, the above results demonstrate the efficacy of the modified ResNet-101 model, (ResNet-101 prob.) with probabilistic reasoning, which addresses the difficulties of multi-label classification on the COCO dataset. High MAP, precision, recall, and F1 Score values demonstrate how well this method works in a variety of categories, enabling its use in extensive, multiple-image classification tasks.

\section{Limitations and future work}
Probabilistic reasoning was implemented into this ResNet-101 model to reduce the dependencies and correlation between different labels, and although in the result (Fig. 2, table II), it is evident that the model is also predicting probabilities of occurring in Dining Table [0.09] and Fork [0.07]. Upon careful observation, it is noted that there is no dining table or fork in Fig. 2. This correlation between the labels is the underlying cause of the observed prediction error. To mitigate this correlation in future work, efforts should be directed towards using Bayesian neural networks and exploring the integration of probabilistic reasoning with other backbone architectures, such as EfficientNet \cite{sovrasov2022combining} or Vision Transformers \cite{ru2022learning}, which might yield better, improved results in the future landscape.

The suggested model's scalability is still another drawback. The model works well on the COCO-2014 dataset, but it hasn't been tried on bigger datasets with more varied and unbalanced labels. To validate the model's robustness and make it feasible to implement in real-world circumstances, more testing with more recent datasets would be necessary.

Because of the enormous number of labels with an unbalanced distribution in the COCO-2014 dataset, probabilistic layers may have had trouble with underrepresented labels, which would have resulted in subpar performance on less frequent classes. Future research should address this problem by improving the prediction accuracy of underrepresented classes in the COCO-2014 dataset using methods such as focal loss and data enhancement.

\section{Conclusions}
Using the COCO-2014 dataset, the work demonstrated a multilabel image classification model in this proposed study that improves label prediction accuracy by conjugating ResNet-101 with probabilistic reasoning layers. As demonstrated by the findings (Fig. 2, Table II), the suggested design improves multilabel classification performance even on overlapped images by efficiently capturing complicated label dependencies and uncertainties, thanks to probabilistic reasoning layers.
Despite the encouraging outcomes, several drawbacks were noted. More research is still needed to fully understand the ResNet-101 architecture's high computing cost, as well as the model's performance on underrepresented classes and generalization to other datasets. The use of lightweight architectures like MobileNet or EfficientNet is suggested for future work, which could lower processing costs without sacrificing accuracy. Additionally, by employing Bayesian neural networks and data augmentation, underrepresented classes can be improved.

\bibliographystyle{unsrt}
\bibliography{references}

\end{document}